%% file: iclr2027_conference.tex
\documentclass{article} 
\usepackage{iclr2027_conference, time}

\iclrfinalcopy

\input{math_commands.tex}

\usepackage{hyperref}
\usepackage{url}
\usepackage{graphicx} 
\usepackage{xcolor}
\usepackage{booktabs}
\usepackage{amsmath}
\usepackage{multirow}
\usepackage{array}
\usepackage{subcaption}
\usepackage{placeins}

\title{Representation by Design in Generation: \\
Cross-View Class-Token Alignment in Diffusion Transformers}

\author{%
Xiaoyu Wu$^{1}$ \quad
Yifei Wang$^{2}$ \quad
Chen Wei$^{2}$ \\[0.4em]
$^{1}$Carnegie Mellon University
\qquad
$^{2}$Rice University \\
\texttt{xiaoyuwu@andrew.cmu.edu, yw251@rice.edu,  cw220@rice.edu} \\
}

\begin{document}

\maketitle
\lhead{Under Review}

\begin{abstract}
Generative and representation learning remain asymmetrically connected:
semantic representations are used to improve diffusion generation,
whereas the models' own representations are often treated as a by-product
of synthesis. We ask whether diffusion models can instead be trained
to learn substantially stronger semantic representations without
sacrificing generation quality.
SelfFlow takes a step in this direction by introducing self-supervised
patch alignment into flow matching, but its main gains remain in faster
convergence and improved generation. Inspired by DINO and iBOT, we
extend this framework with cross-view class-token alignment to further
strengthen semantic representations.
Specifically, we form two independently noised, dual-timestep observations
of each image and align each student class-token representation with
the stop-gradient EMA-teacher target from the other observation.
This objective is optimized jointly with the inherited flow-matching
and local patch objectives. Notably, although the additional objective acts only on the class token,
it strengthens both class-token and patch representations.
Compared with a matched two-view baseline, ImageNet linear-probing
accuracy improves by 9.4\% using the class token and 10.1\%
using mean-pooled patch tokens, while frozen-backbone VOC2012
segmentation improves by 3.6 mIoU.
These representation gains are achieved while maintaining comparable
ImageNet generation FID. In text-to-image training, the same objective
also improves generation FID, reducing it from 2.52 to 2.37 at matched
checkpoints.
Our results show that representation need not remain a by-product of
generation or merely a tool for improving it: it can be directly
optimized as a first-class capability of diffusion pretraining
alongside generation.
\end{abstract}

\section{Introduction}
\label{sec:introduction}

Generative and representation learning have become increasingly connected,
but not symmetrically. Semantic representations are routinely introduced
into diffusion models to accelerate training or improve generation,
whether through feature alignment, semantic latent modeling, or
representation-based autoencoders~\citep{
    yu2025repa,
    wu2025reg,
    kouzelis2025redi,
    zheng2025rae}.
In the other direction, pretrained diffusion models are commonly probed
or adapted for correspondence, segmentation, and recognition~\citep{
    baranchuk2022ddpmseg,
    luo2023diffusionhyperfeatures,
    tang2023dift,
    zhao2023vpd,
    yang2023repfusion},
but their representations are often just treated as a by-product of generation.
The asymmetry is therefore clear: representation learning is often introduced into generative training as a means to improve generation, whereas representation quality itself is rarely treated as an explicit end goal of generative pretraining. This raises a direct question: \emph{can a diffusion model learn substantially stronger semantic representations without sacrificing generation quality?}

Recent methods begin to pursue both goals by incorporating self-supervision
into diffusion training~\citep{
    zhu2024sddit,
    jiang2026sra,
    chefer2026selfflow}.
SelfFlow~\citep{chefer2026selfflow}, in particular, combines flow matching
with alignment to cleaner EMA-teacher patch representations without
relying on an external semantic encoder. This approach improves
convergence and synthesis across modalities, but its reported peak
linear-probing accuracy increases by only about one percentage point
over the flow-matching baseline.

Inspired by DINO's cross-view class-token self-distillation and iBOT's
usage of partial observations~\citep{caron2021dino,zhou2022ibot}, we extend
this framework with cross-view class-token alignment to further
strengthen semantic representations. For each image, we use two
independent Gaussian noise realizations to form two dual-timestep
observations. Each observation contains cleaner tokens that retain
relatively complete image content and noisier tokens that provide
partial information. To encourage a global representation of the
image content shared across these observations, we prepend a CLS token
and align each student class-token representation with the stop-gradient
EMA-teacher target from the other noise branch. This alignment objective
is optimized jointly with the inherited flow-matching and same-view
patch objectives, which remain unchanged.

To ensure a fair comparison, the baseline and our method differ only in
whether cross-view class-token alignment is active, as illustrated in
Figure~\ref{fig:method}. Both use the same transformer architecture,
added CLS token, independently noised dual-timestep inputs, EMA teacher,
flow-matching objective, and local patch objective. Adding this single
objective substantially improves ImageNet linear probing for both CLS
and patch tokens. Patch-token accuracy increases by $10.1\%$,
from $60.0\%$ to $70.1\%$, while frozen-backbone semantic segmentation
on VOC2012 improves by $3.6$ mIoU, from $57.2$ to $60.8$
(Table~\ref{tab:imagenet_main}).
Both the patch-token probe and the segmentation head operate directly
on frozen patch features without using the CLS readout. Thus, although
the additional objective acts only on CLS, these gains reflect stronger
patch representations rather than merely a better global readout.
At the same time, ImageNet generation FID remains comparable to
the baseline. In text-to-image training, the same objective also improves generation
FID, reducing it from 2.52 to 2.37 at matched checkpoints.

Our results suggest that representation
need not remain a by-product of generation or merely a tool for
improving it. Instead, it can be optimized directly alongside synthesis
as a first-class objective of diffusion pretraining, motivating future
work on models that develop semantic and generative capabilities together.
\section{Related Work}
\label{sec:related_work}

\textbf{Semantic representations for diffusion generation.}
A growing line of work introduces semantic representations into diffusion
models to improve optimization and synthesis. REPA aligns intermediate
denoiser features with representations from a frozen self-supervised visual
encoder~\citep{yu2025repa}. REG and ReDi incorporate semantics more directly
into the generative state: REG jointly denoises image latents and a global
representation token, while ReDi jointly models image latents and high-level
visual features~\citep{wu2025reg,kouzelis2025redi}. RAE goes further by
replacing conventional VAE latents with the features of a pretrained
representation encoder~\citep{zheng2025rae}. A parallel line seeks similar
benefits without relying on an external encoder. SD-DiT introduces
self-supervised discrimination into diffusion training, SRA aligns the
model's own representations across layers and noise levels, and Self-Flow
combines internal feature distillation with Dual-Timestep Scheduling for flow
matching~\citep{zhu2024sddit,jiang2026sra,chefer2026selfflow}.
Conditioning Residuals instead feed compact summaries of internal features
back through the denoiser's conditioning
pathway~\citep{xiang2026conditioningresiduals}. Although these approaches
differ in where and how semantics are introduced, representation learning is
primarily used to accelerate generative training or improve synthesis.

\textbf{Diffusion models as representation learners.}
A complementary line studies the semantic structure that emerges from
diffusion models trained for generation. Intermediate denoising features have
been extracted for label-efficient semantic segmentation, semantic
correspondence, and dense visual
perception~\citep{baranchuk2022ddpmseg,tang2023dift,luo2023diffusionhyperfeatures,zhao2023vpd}. DDAE systematically evaluates the linearly separable features learned
through unconditional diffusion pretraining, while DiffMAE modifies diffusion
training with masked observations to prioritize downstream
recognition~\citep{xiang2023ddae,wei2023diffmae}. More closely related to joint
learning, D$^3$CL treats different diffusion noise levels as contrastive
views when adapting a pretrained Stable Diffusion
model~\citep{dai2026probing}. Together, these studies establish
that diffusion models can support visual understanding, but the two directions
remain uneven: semantic representations are commonly optimized when they
benefit generation, whereas representations from generative models are more
often probed, adapted, or transferred after generative pretraining. Our work
focuses on the less explored setting in which semantic quality is itself a
primary objective of diffusion pretraining and is evaluated alongside
generation under a matched generative setup.

\section{Method}
\label{sec:method}

\begin{figure}[t]
   
        \centering
        \includegraphics[width=\linewidth]{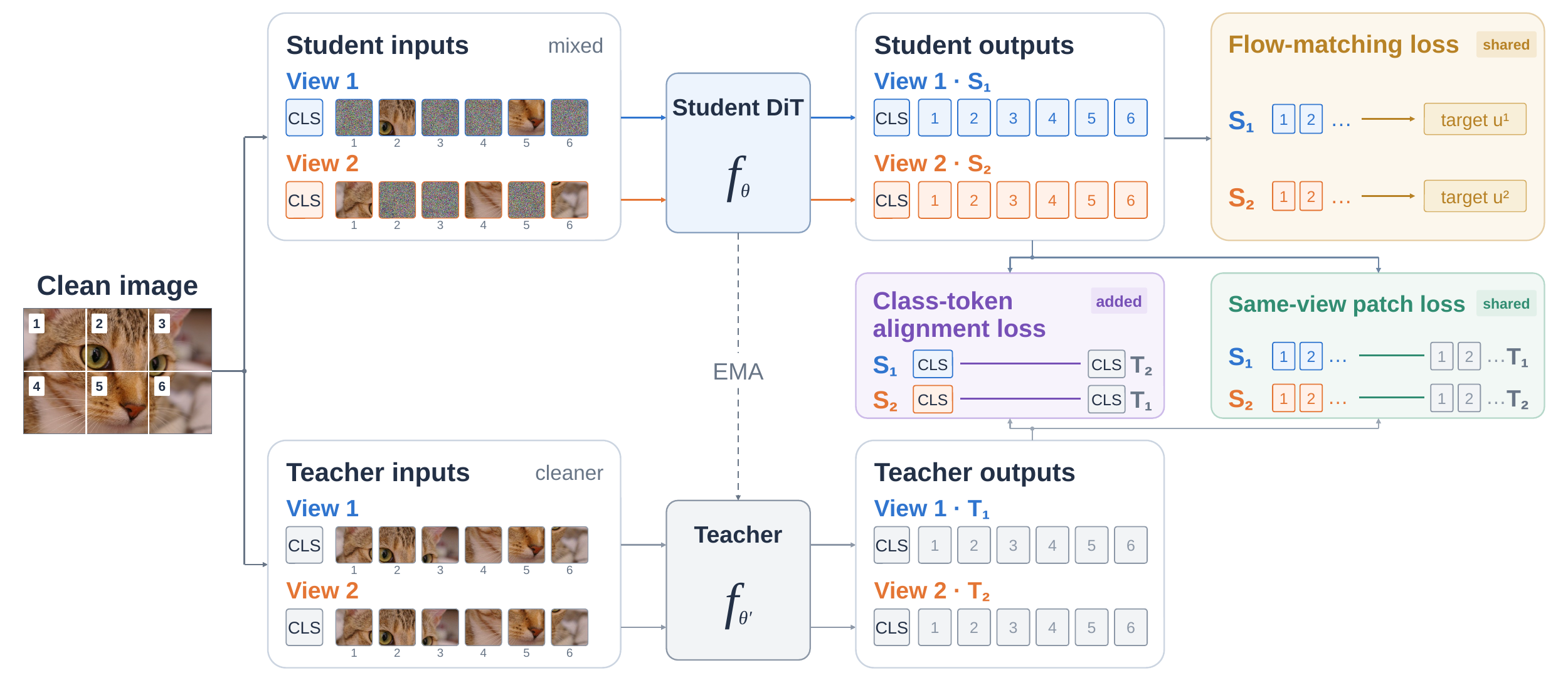}
        \caption{
        \textbf{Class-token alignment across diffusion views.}
        For each clean latent $\mathbf{z}$, two independent noise realizations
        $\boldsymbol{\epsilon}^{(1)}$ and $\boldsymbol{\epsilon}^{(2)}$
        produce mixed-timestep student inputs and corresponding uniformly
        cleaner teacher inputs. The student $f_{\theta}$ and its
        exponential-moving-average (EMA) teacher $f_{\theta'}$ each process
        both views; $S_v$ and $T_v$ label their outputs for view $v$.
        Both variants share flow matching to the targets
        $\mathbf{u}^{(v)}=\mathbf{z}-\boldsymbol{\epsilon}^{(v)}$ and
        same-view patch alignment between $(S_1,T_1)$ and $(S_2,T_2)$.
        Our method additionally aligns class tokens across $(S_1,T_2)$ and
        $(S_2,T_1)$; the matched baseline sets $\lambda_{\mathrm{cls}}=0$.
        Numbered tokens indicate corresponding spatial patches.
        Teacher feature targets are stop-gradient, and the dashed arrow
        denotes the EMA update. The output boxes are schematic: alignment
        features may come from different blocks, and student projection
        heads are omitted.
        }
        \label{fig:method}
   \vspace{-0.1in}

\end{figure}

Our goal is to introduce an explicit global semantic objective into
SelfFlow~\citep{chefer2026selfflow} while preserving its original generative
and local representation objectives. Figure~\ref{fig:method} summarizes the
training pipeline. We first instantiate the dual-timestep process twice using
independent noise, giving two observations of the same image. Both the
baseline and our method use these observations, the same learnable class
(CLS) token and EMA teacher, and the same flow and patch-level objectives.
Our method adds only a cross-view alignment objective on the class token.

\subsection{Two-View SelfFlow Baseline}
\label{sec:method_baseline}

Let $\mathbf{z}\in\mathbb{R}^{N\times d}$ denote a clean image latent
represented by $N$ patch tokens, where $d$ is the latent dimension per
patch. For a noise tensor $\boldsymbol{\epsilon}\in\mathbb{R}^{N\times d}$
with independent standard Gaussian entries, we use the linear transport path
\begin{equation}
    q_t(\mathbf{z},\boldsymbol{\epsilon})
    =
    t\mathbf{z}
    +
    (1-t)\boldsymbol{\epsilon},
    \qquad
    t\in[0,1],
    \label{eq:interpolation}
\end{equation}
where larger $t$ corresponds to a cleaner observation. The velocity along
this path is constant:
\[
    \frac{\partial}{\partial t}
    q_t(\mathbf{z},\boldsymbol{\epsilon})
    =
    \mathbf{z}-\boldsymbol{\epsilon}.
\]

For each training example, we sample
$t_a,t_b\sim\mathcal{U}(0,1)$ and define
\begin{equation}
    t_{\mathrm{c}}=\max(t_a,t_b),
    \qquad
    t_{\mathrm{n}}=\min(t_a,t_b),
\end{equation}
where $t_{\mathrm{c}}$ and $t_{\mathrm{n}}$ denote the cleaner and noisier
timesteps, respectively. The same timestep pair is used for both views,
while their Gaussian noise realizations and token masks are sampled
independently.

For each view $v\in\{1,2\}$, we sample
$\boldsymbol{\epsilon}^{(v)}
\sim\mathcal{N}(\mathbf{0},\mathbf{I})$
and independently draw
$m_i^{(v)}\sim\operatorname{Bernoulli}(\rho)$
for each patch index $i=1,\ldots,N$. Here, $\rho$ is the probability of
selecting the cleaner state ($m_i^{(v)}=1$). The mixed student input
$\widetilde{\mathbf{z}}^{(v)}$ and its per-patch timesteps $\tau_i^{(v)}$
are given by
\begin{align}
    \widetilde{\mathbf{z}}^{(v)}_i
    &=
    m_i^{(v)}
    q_{t_{\mathrm{c}}}
    \left(
        \mathbf{z}_i,
        \boldsymbol{\epsilon}^{(v)}_i
    \right)
    +
    \left(1-m_i^{(v)}\right)
    q_{t_{\mathrm{n}}}
    \left(
        \mathbf{z}_i,
        \boldsymbol{\epsilon}^{(v)}_i
    \right),
    \label{eq:mixed_input}
    \\
    \tau_i^{(v)}
    &=
    m_i^{(v)}t_{\mathrm{c}}
    +
    \left(1-m_i^{(v)}\right)t_{\mathrm{n}}.
    \label{eq:mixed_timestep}
\end{align}
We collect the per-patch timesteps into
$\boldsymbol{\tau}^{(v)}=(\tau_1^{(v)},\ldots,\tau_N^{(v)})$.

The student therefore receives a tokenwise mixture of relatively complete and
partial information. The EMA teacher receives the uniformly cleaner
observation from the corresponding noise branch:
\begin{equation}
    \overline{\mathbf{z}}^{(v)}
    =
    q_{t_{\mathrm{c}}}
    \left(
        \mathbf{z},
        \boldsymbol{\epsilon}^{(v)}
    \right).
    \label{eq:teacher_input}
\end{equation}

Let $f_{\theta}$ denote the student diffusion transformer (DiT) and
$f_{\theta'}$ its EMA teacher, with parameters $\theta$ and $\theta'$,
respectively. Each network is shared across the two views. In
Figure~\ref{fig:method}, $S_v$ and $T_v$ identify the student and teacher
branches for view $v$. We write $f_{\theta}^{\mathrm{flow}}$ for the
student with its velocity-prediction head, giving
\begin{equation}
    \widehat{\mathbf{U}}^{(v)}
    =
    f_{\theta}^{\mathrm{flow}}
    \left(
        \widetilde{\mathbf{z}}^{(v)},
        \boldsymbol{\tau}^{(v)},
        \mathbf{y}
    \right)
    \in\mathbb{R}^{N\times d}
    \label{eq:flow_prediction}
\end{equation}
where $\widehat{\mathbf{U}}^{(v)}$ is the patch-wise velocity prediction
for view $v$ and $\mathbf{y}$ is the class or text condition. The CLS-token
output is not included in $\widehat{\mathbf{U}}^{(v)}$. We denote the
corresponding velocity target by
$\mathbf{u}^{(v)}=\mathbf{z}-\boldsymbol{\epsilon}^{(v)}$, shown as $u^1$
and $u^2$ in Figure~\ref{fig:method}. We optimize
\begin{equation}
    \mathcal{L}_{\mathrm{flow}}
    =
    \frac{1}{2}
    \sum_{v=1}^{2}
    \operatorname{MSE}
    \left(
        \widehat{\mathbf{U}}^{(v)},
        \mathbf{u}^{(v)}
    \right),
    \label{eq:flow_loss}
\end{equation}
where $\operatorname{MSE}$ averages over patch and feature dimensions.
Although the velocity target is constant along this linear path, the
per-token timestep vector $\boldsymbol{\tau}^{(v)}$ remains an input to the
transformer because it specifies the noise state of each patch.

As in SelfFlow, the baseline additionally aligns student patch features with
cleaner EMA-teacher patch features. Let
$\mathbf{P}^{(v)}_{\theta,\ell^{p}_{s}}$ denote student patch-token hidden
features from block $\ell^{p}_{s}$ and
$\mathbf{P}^{(v)}_{\theta',\ell^{p}_{t}}$ denote EMA-teacher patch-token
hidden features from block $\ell^{p}_{t}$. The subscript $i$ indexes
corresponding spatial patches. The local objective is:
\begin{equation}
    \mathcal{L}_{\mathrm{patch}}
    =
    -\frac{1}{2N}
    \sum_{v=1}^{2}
    \sum_{i=1}^{N}
    \operatorname{cos}
    \left(
        g_{\mathrm{patch}}
        \left(
            \mathbf{P}^{(v)}_{\theta,\ell^{p}_{s},i}
        \right),
        \operatorname{sg}
        \left[
            \mathbf{P}^{(v)}_{\theta',\ell^{p}_{t},i}
        \right]
    \right),
    \label{eq:patch_loss}
\end{equation}
where $\operatorname{cos}$ denotes cosine similarity,
$g_{\mathrm{patch}}$ is a two-layer student projection head, and
$\operatorname{sg}$ denotes stop-gradient. Patch alignment is
\emph{same-view}: the student and teacher features use the same noise
realization and corresponding spatial positions. These flow and patch
objectives define our two-view SelfFlow baseline.

\subsection{Cross-View Class-Token Alignment}
\label{sec:cls_distillation}

To introduce an explicit global semantic target, we prepend a learnable CLS
token to the image-token sequence. Unlike patch tokens, the CLS token itself
is not corrupted by the diffusion process. Its per-token timestep shift is
set to zero, while it participates in the same transformer self-attention
and global class or text conditioning as the image tokens. The same token
architecture is used in both the baseline and our method.

Let
$\mathbf{c}^{(v)}_{\theta,\ell^{c}}$ and
$\mathbf{c}^{(v)}_{\theta',\ell^{c}}$
denote the student and EMA-teacher class-token representations from block
$\ell^{c}$. Inspired by class-token self-distillation in DINO and
iBOT~\citep{caron2021dino,zhou2022ibot}, we define a cross-view alignment
objective by matching each student representation to the EMA-teacher
representation from the opposite view:
\begin{align}
    \mathcal{L}_{\mathrm{cls}}
    =
    -\frac{1}{2}
    \Big(
    &
    \operatorname{cos}
    \left(
        g_{\mathrm{cls}}
        \left(
            \mathbf{c}^{(1)}_{\theta,\ell^{c}}
        \right),
        \operatorname{sg}
        \left[
            \mathbf{c}^{(2)}_{\theta',\ell^{c}}
        \right]
    \right)
    \nonumber\\
    +
    &
    \operatorname{cos}
    \left(
        g_{\mathrm{cls}}
        \left(
            \mathbf{c}^{(2)}_{\theta,\ell^{c}}
        \right),
        \operatorname{sg}
        \left[
            \mathbf{c}^{(1)}_{\theta',\ell^{c}}
        \right]
    \right)
    \Big),
    \label{eq:cls_loss}
\end{align}
where $g_{\mathrm{cls}}$ is a separate two-layer student projection head.

Unlike the patch objective in Eq.~\ref{eq:patch_loss}, class-token alignment is
cross-view: each student representation is matched to the EMA target from
the other noise branch. The alignment therefore encourages the class token
to capture information shared across independent observations rather than
noise-specific details. This is the only additional objective introduced by
our method; the flow and local patch objectives remain unchanged.

The patch and CLS objectives use separate two-layer student projection heads. The projection heads are shared
across the two student views but not across objectives. Teacher targets are raw
EMA features and receive no projection or gradient. The projection heads are
discarded for generation and downstream frozen-feature evaluation.

\subsection{Overall Objective and Controlled Comparison}
\label{sec:overall_objective}

The complete objective is
\begin{equation}
    \mathcal{L}
    =
    \mathcal{L}_{\mathrm{flow}}
    +
    \lambda_{\mathrm{patch}}
    \mathcal{L}_{\mathrm{patch}}
    +
    \lambda_{\mathrm{cls}}
    \mathcal{L}_{\mathrm{cls}}.
    \label{eq:total_loss}
\end{equation}
Here, $\lambda_{\mathrm{patch}}$ and $\lambda_{\mathrm{cls}}$ weight the
patch and class-token alignment objectives, respectively. The teacher
parameters are updated as an exponential moving average of the student:
\begin{equation}
    \theta'
    \leftarrow
    \mu\theta'
    +
    (1-\mu)\theta.
    \label{eq:ema}
\end{equation}
where $\mu$ is the EMA decay coefficient.

Our baseline sets $\lambda_{\mathrm{cls}}=0$ but otherwise uses the same model,
added CLS token, independently noised observations, flow loss, patch loss, and
EMA teacher. Our method instead sets $\lambda_{\mathrm{cls}}>0$.
This design isolates the effect of adding direct cross-view supervision
to the CLS token from changes in architecture, corruption, or generative
supervision.

The global and local objectives need not read from the same transformer block.
We therefore use separate notation for the CLS depth $\ell^{c}$, the student
patch depth $\ell^{p}_{s}$, and the teacher patch depth $\ell^{p}_{t}$. Their
dataset-specific values are given in Table~\ref{tab:training_configs}.

Several prior works also pursue generative models with explicit representation
capabilities. MAGE unifies image synthesis and representation learning through
discrete masked-token modeling, while DDAE++ augments diffusion training with
contrastive self-distillation and an explicit self-conditioning pathway;
D$^3$CL instead contrastively adapts a pretrained Stable Diffusion model across
noise levels~\citep{li2023mage,xiang2025ddae++,dai2026probing}. Our setting is complementary: we keep the generative and local patch
objectives fixed, and the matched variants differ only in whether the added
CLS token receives cross-view self-distillation. This isolates the effect of
an explicit global semantic objective during flow pretraining, without
changing the generative formulation or introducing a separate
representation-feedback pathway.

\section{Experiments}
\label{sec:experiments}

We evaluate whether cross-view class-token alignment strengthens semantic
representations while maintaining generation quality. We first report the
matched ImageNet comparison and text-to-image generation results, then
analyze the learned representations and the choice of CLS alignment depth
and loss weight.

\subsection{Experimental Setup}
\label{sec:experimental_setup}

\begin{table}[t]
    \centering
    \caption{
    \textbf{Training configurations.}
    Layer numbers are one-based transformer-block indices.
    The baseline uses the same settings with
    $\lambda_{\mathrm{cls}}=0$.
    }
    \label{tab:training_configs}
    \resizebox{\linewidth}{!}{
    \begin{tabular}{lcc}
        \toprule
        Setting
        & ImageNet
        & Text-to-image \\
        \midrule
        Backbone
        & SelfFlowDiT-XL/2, 28 blocks
        & Flux-style DiT, 21 feature layers \\
        Training data
        & ImageNet-1K
        & 1M text--image pairs \\
        Resolution
        & $256\times256$
        & $256\times256$ \\
        Reported training step
        & 1M
        & 400K \\
        Global batch size
        & 256
        & 512 \\
        Dual-timestep mask ratio $\rho$
        & 0.50
        & 0.75 \\
        Patch student $\rightarrow$ teacher depth
        & $8\rightarrow20$
        & $6\rightarrow15$ \\
        CLS student $\rightarrow$ teacher depth
        & $18\rightarrow18$
        & $15\rightarrow15$ \\
        $\lambda_{\mathrm{patch}}$
        & 0.8
        & 0.8 \\
        $\lambda_{\mathrm{cls}}$
        & 0.2
        & 2.0 \\
        EMA decay $\mu$
        & 0.9999
        & 0.9999 \\
        \bottomrule
    \end{tabular}
    }
\end{table}

\paragraph{ImageNet training.}
We train class-conditional SelfFlowDiT-XL/2 models on ImageNet-1K at
$256\times256$ resolution~\citep{deng2009imagenet}. Images are encoded using
the Stable Diffusion VAE, producing a $16\times16$ latent patch grid. Both the
baseline and our method are trained for 1M steps and evaluated using EMA
weights. Full optimizer, preprocessing, and sampling details are provided in
Appendix~\ref{app:imagenet_details}.

\paragraph{Text-to-image training.}
To test whether the objective extends beyond class-conditional generation, we
apply it to a Flux-style text-conditioned diffusion
transformer trained on 1M text--image pairs. The CLS and patch objectives are
applied only to the image stream; the text-conditioning pipeline is unchanged.
We compare both variants at 400K training steps. Full architecture and evaluation details are provided in
Appendix~\ref{app:t2i_details}.

\paragraph{Representation evaluation.}
We evaluate frozen global and patch representations with linear classifiers.
For the global probe, we extract the raw CLS token from block 18. For the patch
probe, we extract raw patch tokens from block 20 and average them spatially.
For each probe, we standardize every feature dimension using training-set
statistics before fitting the classifier. Following diffusion-representation
evaluation~\citep{xiang2023ddae}, features are extracted from noised latents at
a fixed evaluation timestep, using the null class condition so that no
ground-truth label is provided to the diffusion backbone.

For dense evaluation, we freeze the backbone and train a per-token linear
segmentation head on Pascal VOC2012~\citep{everingham2010voc}. The head applies
LayerNorm followed by a linear classifier to the raw patch grid and bilinearly
upsamples its predictions to the input resolution. We report mean
intersection-over-union (mIoU). Probe optimization and hyperparameter search
are described in Appendix~\ref{app:probe_protocols}.

\paragraph{Generation evaluation.}
For ImageNet, we report FID from 50,000 EMA samples against the standard
ImageNet-256 ADM reference~\citep{heusel2017fid}. For text-to-image generation,
we generate one sample for each of 50,000 held-out prompts and compute FID
against the corresponding validation images and CLIP score against the
prompts~\citep{radford2021clip}. All text-to-image comparisons use the same
sampler, reference set, feature extractor, and preprocessing. These FID values
are intended for controlled comparison between our two variants rather than
for direct comparison with results obtained using a different evaluator.

\begin{table}[t]
    \centering
\caption{
\textbf{ImageNet representation and generation at 1M steps.}
Patch linear probing and VOC segmentation use features from transformer
block 20. Both models contain the same CLS token and differ only in whether
its cross-view alignment objective is active.
}
    \label{tab:imagenet_main}
    \begin{tabular}{lcccc}
        \toprule
        Method
        & FID $\downarrow$
        & [CLS] linear $\uparrow$
        & Patch linear $\uparrow$
        & VOC mIoU $\uparrow$ \\
        \midrule
        Two-view SelfFlow
        & \textbf{5.12}
        & 63.49
        & 60.03
        & 57.15 \\
+ Class-token alignment
        & 5.26
        & \textbf{72.90}
        & \textbf{70.11}
        & \textbf{60.79} \\
        \midrule
        Difference
        & $+0.14$
        & $+9.41$
        & $+10.08$
        & $+3.64$ \\
        \bottomrule
    \end{tabular}
\end{table}

\subsection{ImageNet Results}
\label{sec:imagenet_main}

Table~\ref{tab:imagenet_main} shows that the additional class-token alignment
objective substantially improves both global and local representations.
Class-token linear probing increases by $9.41$ points, from $63.49\%$ to
$72.90\%$. Mean-patch linear probing improves by $10.08$ points, from
$60.03\%$ to $70.11\%$, and frozen VOC2012 segmentation improves by
$3.64$ mIoU, from $57.15$ to $60.79$.

The patch and dense improvements cannot be explained by a better global
readout alone: both evaluations operate directly on the frozen patch tokens.
Although the only \emph{additional} semantic objective is applied to the CLS
token, it changes the representations learned by the image-token stream under
the otherwise unchanged flow and patch objectives.

\begin{figure}[t]
    \centering

    \includegraphics[width=\linewidth]{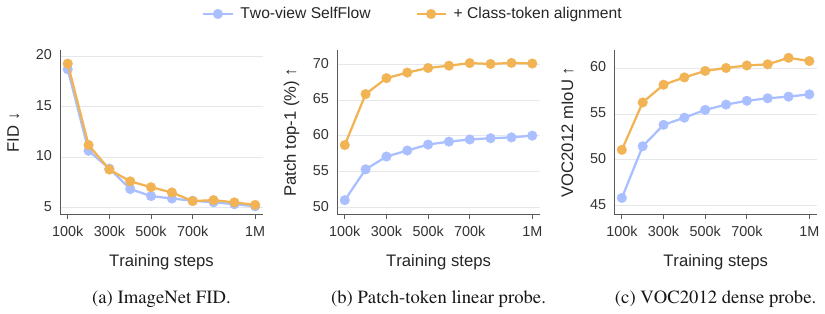}

\caption{
\textbf{ImageNet generation and representation quality throughout training.}
We compare the matched two-view SelfFlow baseline with our model using
cross-view class-token alignment. From left to right, we report ImageNet FID,
patch-token linear-probe accuracy, and dense-probe mIoU across training
checkpoints. Class-token alignment yields persistent improvements in both
patch-level and dense representations while maintaining comparable generation
quality.
}
    \label{fig:imagenet_dynamics}
\end{figure}

These representation gains are achieved with comparable ImageNet generation
FID, which changes from $5.12$ to $5.26$ at the 1M-step checkpoint.
Figure~\ref{fig:imagenet_dynamics} further compares FID, patch-token linear
probing, and VOC2012 mIoU across training checkpoints. The patch and dense
representation gains develop during training, while generation FID follows
a similar trajectory for the two models.

\subsection{Text-to-Image Results}
\label{sec:t2i_results}

We next evaluate whether the same class-token alignment objective also
benefits text-conditioned generation. The baseline and our model use
the same architecture, training data, independently noised observations,
patch objective, and generative objective. They differ only in whether
cross-view class-token alignment is active.

\begin{figure}[t]
    \centering
\includegraphics[width=\linewidth]{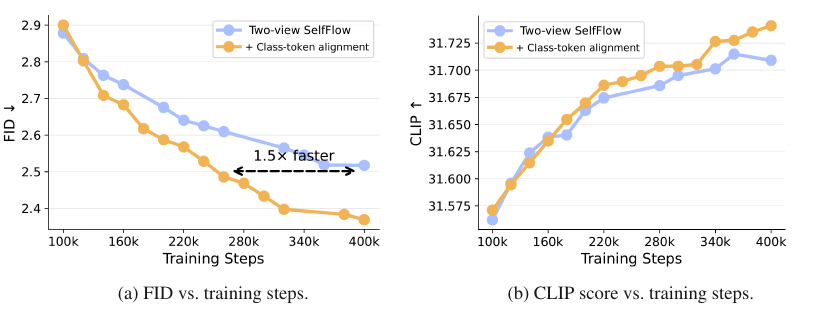}
    \caption{
    \textbf{Text-to-image generation throughout training.}
    We compare the matched two-view SelfFlow baseline with our model using
    cross-view class-token alignment. FID (left) and CLIP score (right) are
    evaluated across training checkpoints using the same sampling and
    evaluation pipeline.
    }
    \label{fig:t2i_training_curves}
\end{figure}

At the matched 400K checkpoint, the additional objective reduces FID from
$2.517$ to $2.369$, while improving CLIP score from $31.709$ to
$31.741$. The training curves in
Figure~\ref{fig:t2i_training_curves} also show faster FID convergence, while
the differences in CLIP score remain small. Thus, the same objective that
strengthens ImageNet representations also benefits generation FID in this
text-to-image setting.

\subsection{Representation Analysis}
\label{sec:representation_analysis}
\label{sec:depth_analysis}
\label{sec:attention_visualization}

\begin{figure}[t]
    \centering

    \includegraphics[width=\linewidth]{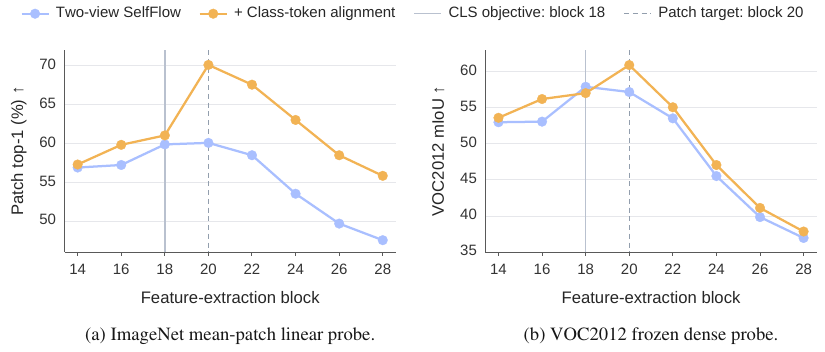}
    \caption{
    \textbf{Patch representation quality across transformer depth.}
    The vertical markers indicate the class-token alignment block (18) and the EMA
    patch-target block (20). The largest gains in both ImageNet patch linear
    probing and VOC2012 dense probing appear at block 20 rather than at the
    directly supervised CLS block.
    }
    \label{fig:depth_sweep}
\end{figure}

\paragraph{Feature-extraction depth.}
We first examine where the patch-level gains appear within the trained model.
The 1M-step checkpoints and the CLS alignment block (18) are fixed; only the
block used to extract patch features is varied.
Figure~\ref{fig:depth_sweep} reports frozen patch evaluation across the later
transformer blocks. At block 18, patch linear accuracy improves only from
$59.85\%$ to $61.01\%$, while VOC mIoU changes from $57.91$ to $57.02$.
At block 20, however, patch linear accuracy increases from $60.05\%$ to
$70.08\%$ and VOC mIoU increases from $57.15$ to $60.79$.

Block 20 is also the layer used as the EMA target for the inherited local patch
objective. The strongest patch-level gains therefore appear at the
patch-teacher depth rather than at the directly aligned CLS depth.
Large patch-linear improvements remain visible in subsequent blocks.

\begin{figure}[t]
    \centering
    \includegraphics[width=\linewidth]{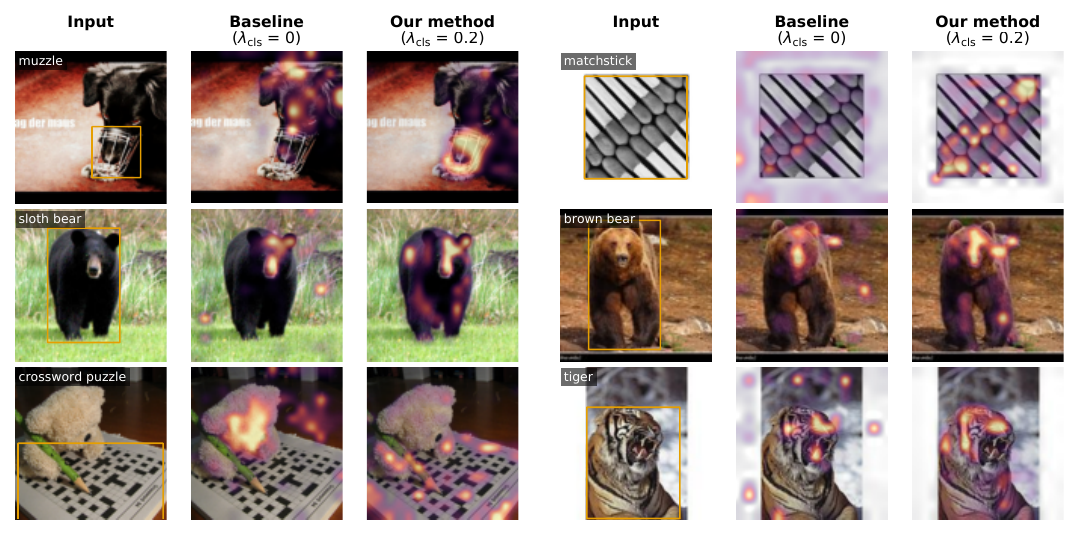}
\caption{
\textbf{CLS-to-patch attention maps on ImageNet.}
Orange boxes indicate the target object regions in the input images. We compare
the matched baseline ($\lambda_{\mathrm{cls}}=0$) with our model using
class-token alignment ($\lambda_{\mathrm{cls}}=0.2$). Across these selected
examples, the aligned model concentrates attention more strongly on
class-relevant and discriminative image regions.
}
    \label{fig:cls_attention}
\end{figure}

\paragraph{CLS-to-patch attention.}

Figure~\ref{fig:cls_attention} compares matched CLS-to-patch attention maps.
In these selected examples, our method concentrates attention more strongly
on class-relevant regions---the muzzle, matchsticks, animal heads, and
crossword grid---than the baseline. These depth comparisons and attention
maps describe the learned representations but do not establish a particular
causal mechanism for the patch-level gains.

\subsection{Ablation Studies}
\label{sec:ablations}

We select the ImageNet CLS attachment depth and loss weight using
100K-step pilot experiments, evaluating representation quality with
weighted $k$NN and generation with FID
(Figure~\ref{fig:cls_ablations}).

\input{figure_insertion.tex}

\paragraph{CLS attachment depth.}
The response is non-monotonic: block 18 gives the highest same-block
patch $k$NN accuracy, block 28 the highest CLS $k$NN accuracy, and
block 20 the lowest FID. We choose block 18 as a trade-off among
these criteria.

\paragraph{CLS objective weight.}
At block 18, CLS $k$NN gains begin to saturate around
$\lambda_{\mathrm{cls}}=0.2$--$0.3$. Increasing the weight from
0.3 to 0.5 worsens FID without improving CLS $k$NN accuracy.
We use $\lambda_{\mathrm{cls}}=0.2$, which has lower FID than 0.3
with slightly lower CLS and patch accuracies.
Exact results are provided in
Tables~\ref{tab:app_cls_depth} and~\ref{tab:app_cls_weight}
in Appendix~\ref{app:cls_ablations}.

\section{Conclusion}
\label{sec:conclusion}
We studied whether semantic representation can be explicitly strengthened during diffusion pretraining without sacrificing generation. Starting from a matched two-view SelfFlow baseline, we add cross-view class-token alignment while retaining the same generative and local patch objectives. This simple addition substantially improves global and patch-token probing and dense semantic segmentation, while preserving ImageNet generation and improving FID and CLIP score in text-to-image training. These results suggest that representation need not remain a by-product of
generation or merely a tool for improving it; it can be optimized directly
alongside synthesis as a first-class outcome of diffusion pretraining.

\newpage

\bibliography{iclr2027_conference}
\bibliographystyle{iclr2027_conference}

\newpage
\appendix

\input{appendix_updated}

\end{document}

%% file: math_commands.tex
\usepackage{amsmath,amsfonts,bm}

\def\eqref#1{equation~\ref{#1}}

\def\1{\bm{1}}

\DeclareMathAlphabet{\mathsfit}{\encodingdefault}{\sfdefault}{m}{sl}
\SetMathAlphabet{\mathsfit}{bold}{\encodingdefault}{\sfdefault}{bx}{n}



%% file: figure_insertion.tex
\begin{figure}[t]
    \centering

    \vspace{-0.2in}
    \includegraphics[width=\linewidth,trim=0 18bp 0 0,clip]{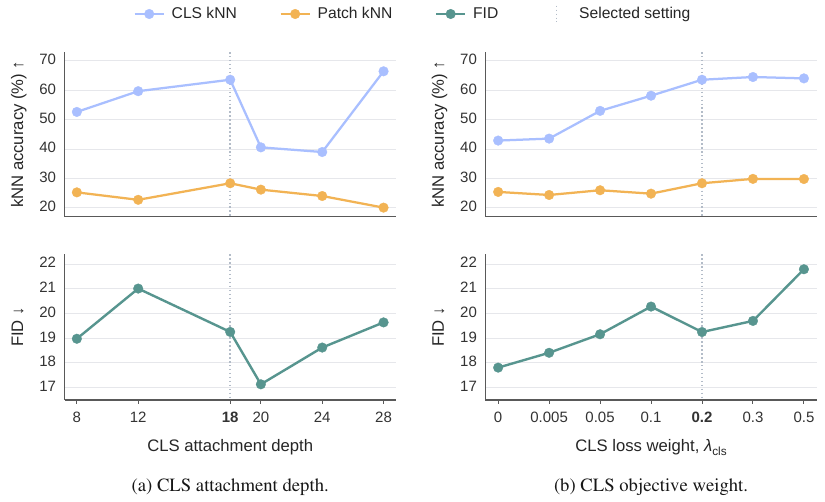}
    \caption{
    \textbf{CLS-objective ablations at 100K ImageNet steps.}
    The attachment-depth sweep evaluates weighted $k$NN at the
    corresponding attachment block; the loss-weight sweep fixes
    the attachment at block 18. Top: CLS and patch $k$NN accuracy.
    Bottom: FID. Dotted guides and bold ticks mark the selected
    settings (block 18 and $\lambda_{\mathrm{cls}}=0.2$).
    Exact values are in Tables~\ref{tab:app_cls_depth}
    and~\ref{tab:app_cls_weight}.
    }
    \label{fig:cls_ablations}
    \vspace{-0.15in}
\end{figure}

%% file: appendix_updated.tex
\section{ImageNet Training and Generation Details}
\label{app:imagenet_details}

\paragraph{Architecture and latent encoding.}
We use a class-conditional SelfFlowDiT-XL/2 backbone with the configuration
in Table~\ref{tab:app_imagenet_config}, initialized from scratch. Images
are resized and center-cropped to $256\times256$, horizontally flipped
with probability $0.5$ during pretraining, and encoded using
\texttt{stabilityai/sd-vae-ft-ema}. Pretraining samples the VAE posterior
and scales the latent by $0.18215$; frozen-feature extraction instead uses
the posterior mode with the same scaling. Each $32\times32\times4$
latent is divided into non-overlapping $2\times2$ patches, giving a
$16\times16$ image-token grid. Both variants prepend a learnable CLS token.

\paragraph{Training and alignment heads.}
Both models are trained for 1M optimizer steps and evaluated using EMA
weights. The global batch contains 256 source images before two-view
expansion, yielding 512 student-view instances per update. Classifier-free
conditioning dropout is sampled independently for the two student
forwards; both EMA-teacher forwards retain the original class labels.
The dropout probability and optimizer settings are listed in
Table~\ref{tab:app_imagenet_config}.

Each active student projection head is
$1152\!\rightarrow\!2304\!\rightarrow\!1152$, with SiLU between the
linear layers. Our method uses separate patch and CLS heads; the baseline
retains the CLS token but does not enable the CLS projection head or
objective. Teacher targets are raw, stop-gradient EMA features. Projection
heads are discarded for generation and frozen-feature evaluation.

For CLS conditioning, we zero the timestep-embedding contribution before
adding class conditioning. This is not equivalent to passing numerical
timestep zero through the timestep embedder. Both ImageNet variants use
this convention. Token masks are sampled independently from a Bernoulli
distribution, with cleaner-token probability $\rho=0.5$ as defined in the
method.

\begin{table}[!htbp]
    \centering
    \caption{\textbf{ImageNet training configuration.}
    Transformer-block indices are one-based. The baseline uses the listed
    backbone and optimization settings with $\lambda_{\mathrm{cls}}=0$
    and no active CLS projection head.}
    \label{tab:app_imagenet_config}
    \begin{tabular}{lc}
        \toprule
        Setting & Value \\
        \midrule
        Dataset & ImageNet-1K \\
        Resolution & $256\times256$ \\
        Backbone & SelfFlowDiT-XL/2 \\
        Transformer blocks & 28 \\
        Hidden dimension & 1152 \\
        Attention heads & 16 \\
        Latent patch grid & $16\times16$ \\
        Global batch (source images) & 256 \\
        Optimizer & AdamW \\
        Learning rate & $10^{-4}$, constant \\
        Adam betas & $(0.9,0.999)$ \\
        Weight decay / warmup steps & $0\,/\,0$ \\
        Training steps & 1M \\
        EMA decay & 0.9999 \\
        Student class-dropout probability & 0.1 \\
        Cleaner-token probability $\rho$ & 0.5 \\
        Patch student $\rightarrow$ teacher block & $8\rightarrow20$ \\
        CLS student/teacher block & 18 \\
        $\lambda_{\mathrm{patch}}$ & 0.8 \\
        $\lambda_{\mathrm{cls}}$ & 0.2 \\
        \bottomrule
    \end{tabular}
\end{table}

\paragraph{Generation evaluation.}
We compute FID~\citep{heusel2017fid} from 50,000 EMA samples using the
ADM/guided-diffusion evaluator and its labeled ImageNet-$256$ reference,
\texttt{VIRTUAL\_\allowbreak imagenet256\_\allowbreak labeled.npz}. Sampling records identify the
adaptive Dopri5 ODE solver with a nominal count of 250. The solver API
interprets this parameter as requested output times, not a verified
number of function evaluations. Classifier-free guidance uses
\begin{equation}
    v_s = v_{\mathrm{uncond}}
        + s\bigl(v_{\mathrm{cond}}-v_{\mathrm{uncond}}\bigr),
    \qquad s=1,
    \label{eq:app_imagenet_cfg}
\end{equation}
so the evaluated setting returns the conditional velocity prediction.
Both variants use the same recorded sampling and evaluation configuration.

\section{Text-to-Image Training and Evaluation Details}
\label{app:t2i_details}

\paragraph{Data and split construction.}
We use the dataset \texttt{nyu-visionx/\allowbreak scale-rae-data},
specifically its \texttt{flux\_\allowbreak training\_\allowbreak 2\_\allowbreak 47m} subset. Shards are processed sequentially,
using seed 42 and a nominal validation-assignment probability of $1/21$,
subject to quotas of 1M training and 50K validation pairs. The resulting
manifests have no overlapping source IDs or prompts under either exact
or whitespace-normalized matching. This check does not test for
duplicates in image content.

\paragraph{Architecture and conditioning.}
We train a Flux-style diffusion transformer from scratch at
$256\times256$ resolution. Its seven double-stream and fourteen
single-stream blocks provide 21 feature layers. The frozen VAE is
\texttt{stabilityai/sd-vae-ft-mse}. A frozen Qwen3-4B text-encoder preset
concatenates features from its configured layers $\{9,18,27\}$ to form
7680-dimensional conditioning features. Prompt dropout is applied with
probability $0.1$, and the resulting context is shared across both
student and both EMA-teacher forwards.

\paragraph{Optimization and objectives.}
The shared data and optimization settings are summarized in
Table~\ref{tab:app_t2i_config}. A global batch contains
$8\times32\times2=512$ source images across eight ranks and two gradient
accumulation steps, before two-view expansion. The patch objective
matches image-stream features from student block 6 to teacher block 15.
Our method additionally applies cross-view CLS alignment at block 15.
Separate runs explored $\lambda_{\mathrm{cls}}\in\{0.5,1,2,4\}$;
the reported 400K-step comparison uses $\lambda_{\mathrm{cls}}=2$.

The text-to-image mask independently selects the noisier branch with
probability $0.25$. Under the method's convention that $\rho$ denotes
cleaner-token probability, this corresponds to $\rho=0.75$.

\begin{table}[!htbp]
    \centering
    \caption{\textbf{Text-to-image training configuration.}
    Diffusion-transformer block indices are one-based. The table lists
    shared architecture and optimization settings, with the method's CLS
    objective; the baseline uses $\lambda_{\mathrm{cls}}=0$.
    Here $\rho$ denotes cleaner-token probability.}
    \label{tab:app_t2i_config}
    \begin{tabular}{lc}
        \toprule
        Setting & Value \\
        \midrule
        Training / validation pairs & 1M / 50K \\
        Resolution & $256\times256$ \\
        Hidden dimension & 1152 \\
        Attention heads & 16 \\
        Double-stream / single-stream blocks & 7 / 14 \\
        Text-conditioning dimension & 7680 \\
        Reported training checkpoint & 400K \\
        Global batch (source images) & 512 \\
        Ranks $\times$ per-rank batch $\times$ accumulation & $8\times32\times2$ \\
        Optimizer & AdamW \\
        Learning rate & $10^{-4}$, constant \\
        Adam betas & $(0.9,0.999)$ \\
        Weight decay / warmup steps & $0\,/\,0$ \\
        Gradient clipping & 1.0 \\
        EMA decay & 0.9999 \\
        Prompt-dropout probability & 0.1 \\
        Cleaner-token probability $\rho$ & 0.75 \\
        Patch student $\rightarrow$ teacher block & $6\rightarrow15$ \\
        CLS student/teacher block & 15 \\
        $\lambda_{\mathrm{patch}}$ & 0.8 \\
        $\lambda_{\mathrm{cls}}$ & 2.0 \\
        \bottomrule
    \end{tabular}
\end{table}

\paragraph{Sampling and FID.}
We generate one image for each of 50,000 held-out prompts using the EMA
checkpoint at $256\times256$ resolution. The reported evaluation settings
are 50 sampling steps and guidance scale 1.0. FID compares generated
images with the corresponding 50,000 validation images.

The text-to-image evaluator uses \texttt{torchvision} Inception V3 with
\texttt{DEFAULT} pretrained weights, \texttt{fc=Identity}, and
\texttt{transform\_input=False}. Images are converted to RGB and bicubically
resized to $256\times256$. The weights' preprocessing then resizes with
bilinear interpolation to size 342 and center-crops to $299\times299$
before feature extraction. We use this same pipeline and reference set
for every reported text-to-image FID. These values are intended for
within-pipeline comparisons, not direct comparison with FID values from
the separate ADM evaluation used for ImageNet or other feature pipelines.

\paragraph{CLIP score.}
We use \texttt{openai/clip-vit-base-patch32}~\citep{radford2021clip}.
For image and text embeddings independently normalized to unit length,
$\hat e_i^{\mathrm{image}}$ and $\hat e_i^{\mathrm{text}}$, the score is
\begin{equation}
    \mathrm{CLIP}
    = \frac{100}{n}\sum_{i=1}^{n}
      \max\!\left\{
      \bigl(\hat e_i^{\mathrm{image}}\bigr)^{\mathsf T}
      \hat e_i^{\mathrm{text}},\,0\right\},
    \qquad n=50{,}000.
    \label{eq:app_clip_score}
\end{equation}
Thus, negative cosine similarities are clamped to zero before averaging.
All comparisons use the same prompts and CLIP preprocessing.

\section{Representation Evaluation}
\label{app:probe_protocols}

\subsection{ImageNet Linear Probing}
\label{app:imagenet_linear}

\paragraph{Feature extraction.}
We evaluate the frozen 1M-step EMA checkpoints on all 1,281,167 ImageNet
training images and all 50,000 validation images. Images are encoded using
the deterministic VAE posterior mode with the $0.18215$ scaling described
in Appendix~\ref{app:imagenet_details}. At evaluation timestep $t=0.8$, the
backbone receives
\begin{equation}
    z_t=0.8z+0.2\epsilon,
    \qquad \epsilon\sim\mathcal{N}(0,I),
    \label{eq:app_eval_noise}
\end{equation}
consistent with the transport convention in the method. The backbone
receives the null class condition, not the ground-truth ImageNet label.

For the global representation, we extract the raw CLS token after block 18.
For the local representation, we spatially average the raw $16\times16$
patch-token grid after block 20. We do not apply samplewise
$\ell_2$ normalization. Instead, each feature dimension is standardized
using its mean and standard deviation on the training set; the same
statistics are applied to validation features.

Each image receives one Gaussian draw during extraction, and its resulting
feature is cached. These features are reused across probe epochs and
learning-rate candidates without resampling diffusion noise. Within each
readout, the two models use extraction seed 0 and matching execution
settings: three GPU ranks for CLS features and four for mean-patch features.
Noise tensors were not retained, so exact per-image matching between models
has not been independently verified. CLS and mean-patch caches are
extracted separately and are not assumed to share noise realizations.

\paragraph{Probe training and selection.}
For every model and readout, we train a single affine classifier using SGD
with momentum $0.9$, zero weight decay, batch size 8192, and 60 epochs.
The learning rate is constant, with no warmup. All four final probes use
the same 33-value learning-rate grid:
\begin{equation*}
\begin{split}
\mathcal{H}_{\mathrm{linear}}=\{&
0.005,\ 0.0075,\ 0.010,\ 0.0125,\ 0.015,\ 0.0175,\ 0.020,\ 0.0225,\\
&0.025,\ 0.030,\ 0.040,\ 0.050,\ 0.060,\ 0.070,\ 0.080,\ 0.090,\\
&0.100,\ 0.120,\ 0.140,\ 0.160,\ 0.180,\ 0.200,\ 0.240,\ 0.280,\\
&0.320,\ 0.360,\ 0.420,\ 0.500,\ 0.600,\ 0.750,\ 0.900,\ 1.100,\ 1.300\}.
\end{split}
\end{equation*}
We independently select the learning rate and epoch that maximize
validation top-1 accuracy for each model and readout. Thus, the search
space and selection rule are identical, but the selected hyperparameters
need not be. Table~\ref{tab:app_linear_selected} gives the selected settings
and final results.

\begin{table}[!htbp]
    \centering
    \caption{\textbf{Selected settings for final ImageNet linear probes.}
    Each probe uses the same 33-rate search and 60-epoch training budget.
    Bold and underline mark the higher and lower accuracy within each
    readout, respectively.}
    \label{tab:app_linear_selected}
    \begin{tabular}{lccc}
        \toprule
        Model & Learning rate & Epoch & Top-1 (\%) \\
        \midrule
        \multicolumn{4}{l}{\emph{CLS, block 18}} \\
        Baseline & 0.12 & 54 & \underline{63.488} \\
        Ours     & 0.10 & 17 & \textbf{72.898} \\
        \midrule
        \multicolumn{4}{l}{\emph{Mean patch, block 20}} \\
        Baseline & 0.18 & 58 & \underline{60.03} \\
        Ours     & 0.12 & 57 & \textbf{70.11} \\
        \bottomrule
    \end{tabular}
\end{table}

\paragraph{Evaluation sweeps.}
The feature-extraction-depth and checkpoint-trend patch probes use 27-rate
and 29-rate grids, respectively. Relative to the final grid above,
\begin{equation*}
\begin{split}
\mathcal{H}_{\mathrm{depth}}
    &=\mathcal{H}_{\mathrm{linear}}\setminus
      \{0.005,\,0.0075,\,0.010,\,0.0125,\,0.0175,\,0.0225\},\\
\mathcal{H}_{\mathrm{trend}}
    &=\mathcal{H}_{\mathrm{depth}}\cup\{1.600,\,2.000\}.
\end{split}
\end{equation*}
Both sweeps use the same 60-epoch budget, batch size 8192, SGD momentum,
zero weight decay, constant learning rates without warmup, training-set
standardization, and best-validation selection described above. At 1M
steps, the depth, trend, and final mean-patch evaluations reuse the same
feature caches; differences between these evaluation series are therefore
not independent diffusion-noise trials. The block-20 patch readout was
chosen from the blockwise evaluation, whose full curve is also reported.

\subsection{Pascal VOC2012 Dense Probing}
\label{app:voc_dense}

\paragraph{Data and preprocessing.}
We evaluate frozen dense representations on Pascal
VOC2012~\citep{everingham2010voc}, using 1,464 training images and 1,449
validation images. When needed, images undergo progressive BOX
downsampling, followed by bicubic resizing of the shorter side to 256 and
a $256\times256$ center crop. Segmentation masks undergo the same geometric
transformations with nearest-neighbor interpolation at every resize,
preserving class IDs and the void label 255.

\paragraph{Features and dense head.}
The frozen backbone receives null ImageNet class ID 1000 during both
training and validation. VOC labels supervise only the segmentation head.
The head applies LayerNorm and a tokenwise linear classifier over 21
classes to the raw $16\times16$ patch-token grid from block 20. Logits are
bilinearly upsampled to the mask resolution with
\texttt{align\_corners=False}. We optimize cross-entropy with ignore index
255 and report mIoU from the 21-class confusion matrix.

Features use $t=0.8$ under Eq.~\ref{eq:app_eval_noise}. Unlike ImageNet
linear probing, this evaluation draws fresh Gaussian noise on every
feature-extraction call, including each validation pass. All learning-rate
heads share the features from a given batch. We do not average predictions
across multiple noise realizations.

\paragraph{Optimization and selection.}
We use Adam with betas $(0.9,0.999)$, zero weight decay, batch size 32,
and cosine learning-rate annealing over 60 epochs without warmup. The
initial learning-rate grid is
\begin{equation*}
\begin{split}
\mathcal{H}_{\mathrm{VOC}}=\{&
0.0002,\ 0.0003,\ 0.0005,\ 0.00075,\ 0.001,\\
&0.0015,\ 0.002,\ 0.003,\ 0.0045,\ 0.006\}.
\end{split}
\end{equation*}
Each model independently selects the learning rate and epoch with the
highest validation mIoU. Both select an initial learning rate of $0.0045$:
the baseline reaches $57.15$ mIoU at epoch 48, and our method reaches
$60.79$ mIoU at epoch 19.

\subsection{Weighted $k$NN Evaluation}
\label{app:knn_protocol}

The 100K-step CLS ablations use weighted $k$NN rather than a trained linear
probe. We construct a labeled feature bank from all 1,281,167 ImageNet
training images and evaluate all 50,000 validation images. Frozen EMA
features are extracted using a $256\times256$ center crop, null
conditioning, $t=0.8$, and one Gaussian noise draw per image. Extraction
uses seed 0, four GPU ranks, and batch size 128 per rank.

We $\ell_2$-normalize the CLS vector or mean-pooled patch vector and retrieve
the $k=20$ nearest neighbors by normalized inner product. For a normalized
query $\hat q$, normalized bank features $\hat f_i$, and bank labels $y_i$,
the prediction is
\begin{equation}
    \hat y(\hat q)=
    \mathop{\arg\max}_{c}
    \sum_{i\in\mathcal{N}_{20}(\hat q)}
    \mathbf{1}\{y_i=c\}
    \exp\!\left(\frac{\hat q^{\mathsf T}\hat f_i}{0.07}\right),
    \label{eq:app_knn_vote}
\end{equation}
where $\mathcal{N}_{20}(\hat q)$ is the retrieved neighbor set. The same
$k$ and temperature are used throughout both ablations. The attachment-depth
sweep reads features at each model's attachment block; the loss-weight
sweep always reads block 18.

\subsection{Single-Block and Multi-Block Features}
\label{app:multiblock}

We additionally concatenate features from blocks $\{14,16,18,20\}$.
For ImageNet, we mean-pool each block separately, concatenate the four
1152-dimensional vectors, and standardize each dimension of the resulting
4608-dimensional vector using training-set statistics. A single affine
classifier is trained with the 27-rate grid
$\mathcal{H}_{\mathrm{depth}}$ and the optimization protocol in
Appendix~\ref{app:imagenet_linear}. The selected learning rates and epochs
are $0.07$/60 for the baseline and $0.06$/56 for our method.

For VOC, block features are concatenated along the raw token-channel
dimension. The head is one $\operatorname{LayerNorm}(4608)$ followed by
$\operatorname{Linear}(4608,21)$. Both models select learning rate $0.003$,
with the baseline and our method selecting epochs 52 and 48, respectively.
The resulting scores are listed in Table~\ref{tab:app_multiblock}.
Concatenation improves the reported absolute VOC mIoU for both models,
while the between-model gain is larger at block 20 alone.

The single-block ImageNet row retains the auxiliary scores reported with
this comparison; the final 33-rate results are listed separately in
Table~\ref{tab:app_linear_selected}.

\begin{table}[!htbp]
    \centering
    \caption{\textbf{Single-block and multi-block patch evaluations.}
    ImageNet reports linear-probe top-1 accuracy; VOC reports dense-probe
    mIoU. $\Delta$ is the difference between ours and baseline, in percentage
    points. Within each dataset, bold and underline mark the highest and
    second-highest value in each numerical column.}
    \label{tab:app_multiblock}
    \begin{tabular}{lccc}
        \toprule
        Feature blocks & Baseline & Ours & $\Delta$ \\
        \midrule
        \multicolumn{4}{l}{\emph{ImageNet}} \\
        $20$ & \underline{60.052} & \underline{70.084} & \textbf{+10.032} \\
        $\{14,16,18,20\}$ & \textbf{62.602} & \textbf{70.516} & \underline{+7.914} \\
        \midrule
        \multicolumn{4}{l}{\emph{VOC2012}} \\
        $20$ & \underline{57.180} & \underline{60.912} & \textbf{+3.732} \\
        $\{14,16,18,20\}$ & \textbf{60.399} & \textbf{63.104} & \underline{+2.705} \\
        \bottomrule
    \end{tabular}
\end{table}

\FloatBarrier
\section{Numerical Results for CLS Ablations}
\label{app:cls_ablations}

Tables~\ref{tab:app_cls_depth} and~\ref{tab:app_cls_weight} provide the
exact values for the main-text CLS ablation figure. All checkpoints are
evaluated at 100K training steps using the weighted $k$NN protocol in
Appendix~\ref{app:knn_protocol}. The depth sweep varies both the loss
attachment and representation readout block, rather than isolating
attachment depth under a fixed readout. The weight sweep fixes both at
block 18; fixed objective weights are given in the table captions.
The main ImageNet configuration uses block 18 and
$\lambda_{\mathrm{cls}}=0.2$ as a trade-off, not as the optimum for every
metric. Bold and underline mark the best and second-best value in each
metric column.

\begin{table}[!htbp]
    \centering
    \caption{\textbf{CLS attachment-depth ablation at 100K steps.}
    $\lambda_{\mathrm{cls}}=0.2$ and $\lambda_{\mathrm{patch}}=0.8$
    throughout. Representation metrics use features from the corresponding
    attachment block. $k$NN accuracies are percentages.}
    \label{tab:app_cls_depth}
    \begin{tabular}{rccc}
        \toprule
        CLS block & FID $\downarrow$ & CLS $k$NN $\uparrow$ & Patch $k$NN $\uparrow$ \\
        \midrule
         8 & 18.978 & 52.638 & 25.268 \\
        12 & 21.008 & 59.696 & 22.732 \\
        18 & 19.254 & \underline{63.592} & \textbf{28.366} \\
        20 & \textbf{17.135} & 40.588 & \underline{26.194} \\
        24 & \underline{18.622} & 38.996 & 24.030 \\
        28 & 19.639 & \textbf{66.458} & 20.076 \\
        \bottomrule
    \end{tabular}
\end{table}

\begin{table}[!htbp]
    \centering
    \caption{\textbf{CLS-loss-weight ablation at block 18 and 100K steps.}
    $\lambda_{\mathrm{patch}}=0.8$ throughout. $k$NN accuracies are percentages.}
    \label{tab:app_cls_weight}
    \begin{tabular}{cccc}
        \toprule
        $\lambda_{\mathrm{cls}}$ & FID $\downarrow$ & CLS $k$NN $\uparrow$ & Patch $k$NN $\uparrow$ \\
        \midrule
        0.000 & \textbf{17.81} & 42.89 & 25.42 \\
        0.005 & \underline{18.41} & 43.56 & 24.38 \\
        0.050 & 19.16 & 53.00 & 25.99 \\
        0.100 & 20.28 & 58.16 & 24.83 \\
        0.200 & 19.25 & 63.59 & 28.37 \\
        0.300 & 19.70 & \textbf{64.52} & \textbf{29.86} \\
        0.500 & 21.79 & \underline{64.05} & \underline{29.81} \\
        \bottomrule
    \end{tabular}
\end{table}

\clearpage
\section{Qualitative Text-to-Image Examples}
\label{app:visualization}

\begin{figure}[!htbp]
    \centering
    \includegraphics[
        height=0.78\textheight,
        width=\linewidth,
        keepaspectratio
    ]{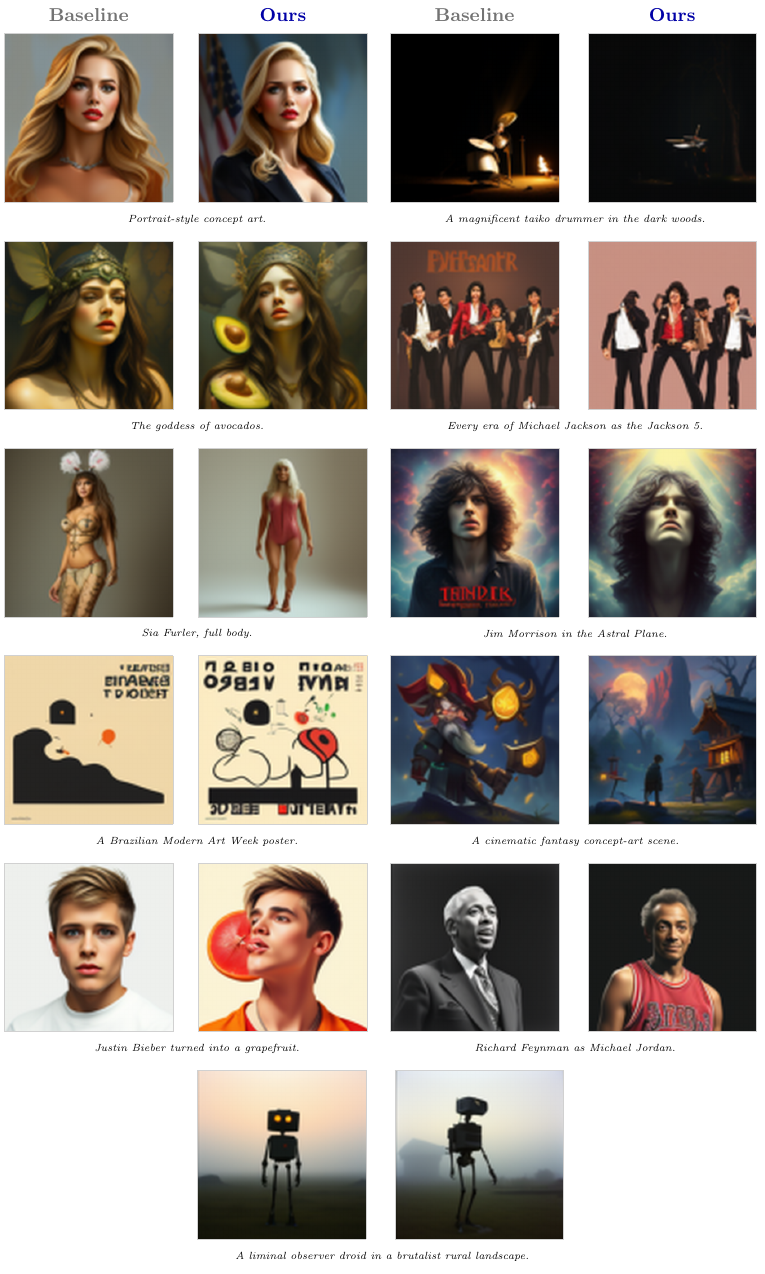}
    \caption{\textbf{Selected text-to-image comparisons at 400K steps.}
    Within each pair, both models use the same prompt, random seed, and
    sampling configuration. Examples were selected post hoc according to
    per-sample CLIP improvement and illustrate individual cases rather than
    average performance. One pair, with prompt ID 3, is omitted because the
    baseline output is nearly blank. Displayed prompt labels are abbreviated;
    full prompts and seeds are listed in
    Table~\ref{tab:t2i_qualitative_prompts}.}
    \label{fig:t2i_qualitative_all}
\end{figure}

\begin{table}[!htbp]
    \centering
    \caption{\textbf{Full prompts and random seeds for
    Figure~\ref{fig:t2i_qualitative_all}.} Original prompt IDs are retained.}
    \label{tab:t2i_qualitative_prompts}
    \begingroup
    \small
    \setlength{\tabcolsep}{4pt}
    \begin{tabular}{@{}c p{0.76\linewidth} r@{}}
        \toprule
        ID & Exact prompt & Seed \\
        \midrule
        1
        & Draw Lindsey Pelas as Gillian Anderson, the president of the United
          States, digital painting, ArtStation concept art, sharp-focus
          illustration art by Artgerm, H 704.
        & 20415 \\
        2
        & I want an image of a magnificent taiko drummer in the dark woods.
        & 17146 \\
        4
        & Make a picture of the goddess of avocados, by Donato Giancola.
        & 49143 \\
        5
        & Produce an image depicting every era of Michael Jackson represented
          as all the members of the Jackson 5.
        & 40777 \\
        6
        & Please visualize Sia Furler full body.
        & 42874 \\
        7
        & I want an image of a festival poster of Jim Morrison in the Astral
          Plane, haunting digital art.
        & 49751 \\
        8
        & Show the Brazilian Modern Art Week poster. The poster style is
          modernism and the details are minimal. The poster features a sequence
          of images, ideas, emotions, and sensations that usually occur
          involuntarily in the mind during certain stages of sleep. The
          background is light beige, modernism, cubism, minimalism, designed by
          George Condo as an image.
        & 22867 \\
        9
        & Produce an image depicting photo cartoon illustration, digital
          painting, volumetric lighting by Feng Zhu, 3D, Alejandro Alvarez
          Alena, Aenami artworks in 4K, Beeple, by Carel Willink and Gregory
          Crewdson, by Thomas Kinkade, Hearthstone, League of Legends, Dofus,
          Overwatch, concept sheet.
        & 34805 \\
        10
        & I'd like to see an image of Justin Bieber turned into a grapefruit.
        & 20273 \\
        11
        & Create a picture that shows Richard Feynman as Michael Jordan.
        & 20203 \\
        12
        & Generate imagery of the lanky liminal observer droid by Dennis
          Mejillones, in a brutalist yet rural landscape by Simon Stalenhag,
          35 mm film photography, dawn, eerie fog.
        & 46176 \\
        \bottomrule
    \end{tabular}
    \endgroup
\end{table}

\FloatBarrier
\section{Limitations}
\label{app:limitations}

Given the computational cost of diffusion pretraining, we prioritize
matched ImageNet comparisons and targeted ablations, with one pretraining
run per configuration in the main experiments. Evaluating variability
across pretraining seeds and extending the study to larger training scales
remain directions for future work.

Our representation evaluation focuses on class-conditional ImageNet
pretraining and dense transfer to Pascal VOC2012. Additional downstream
tasks and frozen-feature evaluation of the text-to-image models would
help assess the generality of the representation gains. Text-to-image
improvements are metric-dependent, with mixed results on compositional
benchmarks. Finally, our depth analyses and attention visualizations
characterize the learned representations; further investigation could
clarify how global alignment strengthens patch features.

%% file: iclr2027_conference.bib
@inproceedings{yu2025repa,
  title={Representation Alignment for Generation: Training Diffusion Transformers Is Easier Than You Think},
  author={Yu, Sihyun and Kwak, Sangkyung and Jang, Huiwon and Jeong, Jongheon and Huang, Jonathan and Shin, Jinwoo and Xie, Saining},
  booktitle={International Conference on Learning Representations},
  editor={Yue, Yisong and Garg, Abhinav and Peng, Nan and Sha, Fei and Yu, Rose},
  pages={87400--87442},
  volume={2025},
  year={2025},
  url={https://proceedings.iclr.cc/paper_files/paper/2025/file/d9e42b4d7163931f3689d6d6fbaa11d0-Paper-Conference.pdf}
}

@inproceedings{wu2025reg,
  title={Representation Entanglement for Generation: Training Diffusion Transformers Is Much Easier Than You Think},
  author={Wu, Ge and Zhang, Shen and Shi, Ruijing and Gao, Shanghua and Chen, Zhenyuan and Wang, Lei and Chen, Zhaowei and Gao, Hongcheng and Tang, Yao and Yang, Jian and Cheng, Ming-Ming and Li, Xiang},
  booktitle={Advances in Neural Information Processing Systems},
  editor={Belgrave, Danielle and Zhang, Cheng and Lin, Haoping and Pascanu, Razvan and Koniusz, Piotr and Ghassemi, Marzyeh and Chen, Nan},
  pages={7714--7743},
  publisher={Curran Associates, Inc.},
  volume={38, Main Conference},
  year={2025},
  doi={10.52202/085713-0264},
  url={https://proceedings.neurips.cc/paper_files/paper/2025/file/0b99315234cc95e6ef281f9155b68832-Paper-Conference.pdf}
}

@inproceedings{kouzelis2025redi,
  title={Boosting Generative Image Modeling via Joint Image-Feature Synthesis},
  author={Kouzelis, Theodoros and Karypidis, Efstathios and Kakogeorgiou, Ioannis and Gidaris, Spyridon and Komodakis, Nikos},
  booktitle={Advances in Neural Information Processing Systems},
  editor={Belgrave, Danielle and Zhang, Cheng and Lin, Haoping and Pascanu, Razvan and Koniusz, Piotr and Ghassemi, Marzyeh and Chen, Nan},
  pages={16685--16714},
  publisher={Curran Associates, Inc.},
  volume={38, Main Conference},
  year={2025},
  doi={10.52202/085713-0563},
  url={https://proceedings.neurips.cc/paper_files/paper/2025/file/186a213d720568b31f9b59c085a23e5a-Paper-Conference.pdf}
}

@inproceedings{zheng2025rae,
  title={Diffusion Transformers with Representation Autoencoders},
  author={Zheng, Boyang and Ma, Nanye and Tong, Shengbang and Xie, Saining},
  booktitle={International Conference on Learning Representations},
  editor={Vondrick, Carl and Hariharan, Bharath and Raffel, Colin and Pinto, Lerrel and Yang, Diyi and Faust, Aleksandra},
  pages={35791--35820},
  volume={2026},
  year={2026},
  url={https://proceedings.iclr.cc/paper_files/paper/2026/file/3c4141c12660ad3625eb4ae845e0a6f9-Paper-Conference.pdf}
}

@inproceedings{baranchuk2022ddpmseg,
  title={Label-Efficient Semantic Segmentation with Diffusion Models},
  author={Baranchuk, Dmitry and Rubachev, Ivan and Voynov, Andrey and Khrulkov, Valentin and Babenko, Artem},
  booktitle={International Conference on Learning Representations},
  year={2022},
  url={https://openreview.net/forum?id=SlxSY2UZQT}
}

@article{luo2023diffusionhyperfeatures,
  title={Diffusion hyperfeatures: Searching through time and space for semantic correspondence},
  author={Luo, Grace and Dunlap, Lisa and Park, Dong Huk and Holynski, Aleksander and Darrell, Trevor},
  journal={Advances in Neural Information Processing Systems},
  volume={36},
  pages={47500--47510},
  year={2023}
}

@article{tang2023dift,
  title={Emergent correspondence from image diffusion},
  author={Tang, Luming and Jia, Menglin and Wang, Qianqian and Phoo, Cheng Perng and Hariharan, Bharath},
  journal={Advances in neural information processing systems},
  volume={36},
  pages={1363--1389},
  year={2023}
}

@inproceedings{zhao2023vpd,
  title={Unleashing text-to-image diffusion models for visual perception},
  author={Zhao, Wenliang and Rao, Yongming and Liu, Zuyan and Liu, Benlin and Zhou, Jie and Lu, Jiwen},
  booktitle={Proceedings of the IEEE/CVF international conference on computer vision},
  pages={5729--5739},
  year={2023}
}

@inproceedings{yang2023repfusion,
  title={Diffusion model as representation learner},
  author={Yang, Xingyi and Wang, Xinchao},
  booktitle={Proceedings of the IEEE/CVF International Conference on Computer Vision},
  pages={18938--18949},
  year={2023}
}

@inproceedings{wei2023diffmae,
  title={Diffusion models as masked autoencoders},
  author={Wei, Chen and Mangalam, Karttikeya and Huang, Po-Yao and Li, Yanghao and Fan, Haoqi and Xu, Hu and Wang, Huiyu and Xie, Cihang and Yuille, Alan and Feichtenhofer, Christoph},
  booktitle={Proceedings of the IEEE/CVF International Conference on Computer Vision},
  pages={16284--16294},
  year={2023}
}

@inproceedings{zhu2024sddit,
  title={{SD-DiT}: Unleashing the Power of Self-supervised Discrimination in Diffusion Transformer},
  author={Zhu, Rui and Pan, Yingwei and Li, Yehao and Yao, Ting and Sun, Zhenglong and Mei, Tao and Chen, Chang Wen},
  booktitle={Proceedings of the IEEE/CVF Conference on Computer Vision and Pattern Recognition (CVPR)},
  month={June},
  year={2024},
  pages={8435--8445}
}

@article{chefer2026selfflow,
  title={Self-supervised flow matching for scalable multi-modal synthesis},
  author={Chefer, Hila and Esser, Patrick and Lorenz, Dominik and Podell, Dustin and Raja, Vikash and Tong, Vinh and Torralba, Antonio and Rombach, Robin},
  journal={arXiv preprint arXiv:2603.06507},
  year={2026}
}

@inproceedings{caron2021dino,
  title={Emerging properties in self-supervised vision transformers},
  author={Caron, Mathilde and Touvron, Hugo and Misra, Ishan and J{\'e}gou, Herv{\'e} and Mairal, Julien and Bojanowski, Piotr and Joulin, Armand},
  booktitle={Proceedings of the IEEE/CVF international conference on computer vision},
  pages={9650--9660},
  year={2021}
}

@article{zhou2022ibot,
  title={ibot: Image bert pre-training with online tokenizer},
  author={Zhou, Jinghao and Wei, Chen and Wang, Huiyu and Shen, Wei and Xie, Cihang and Yuille, Alan and Kong, Tao},
  journal={arXiv preprint arXiv:2111.07832},
  year={2021}
}

@inproceedings{deng2009imagenet,
  title={Imagenet: A large-scale hierarchical image database},
  author={Deng, Jia and Dong, Wei and Socher, Richard and Li, Li-Jia and Li, Kai and Fei-Fei, Li},
  booktitle={2009 IEEE conference on computer vision and pattern recognition},
  pages={248--255},
  year={2009},
  organization={Ieee}
}

@inproceedings{xiang2023ddae,
  title={Denoising diffusion autoencoders are unified self-supervised learners},
  author={Xiang, Weilai and Yang, Hongyu and Huang, Di and Wang, Yunhong},
  booktitle={Proceedings of the IEEE/CVF International Conference on Computer Vision},
  pages={15802--15812},
  year={2023}
}

@article{everingham2010voc,
  title={The pascal visual object classes (voc) challenge},
  author={Everingham, Mark and Van Gool, Luc and Williams, Christopher KI and Winn, John and Zisserman, Andrew},
  journal={International journal of computer vision},
  volume={88},
  number={2},
  pages={303--338},
  year={2010},
  publisher={Springer}
}

@article{heusel2017fid,
  title={Gans trained by a two time-scale update rule converge to a local nash equilibrium},
  author={Heusel, Martin and Ramsauer, Hubert and Unterthiner, Thomas and Nessler, Bernhard and Hochreiter, Sepp},
  journal={Advances in neural information processing systems},
  volume={30},
  year={2017}
}

@inproceedings{radford2021clip,
  title={Learning transferable visual models from natural language supervision},
  author={Radford, Alec and Kim, Jong Wook and Hallacy, Chris and Ramesh, Aditya and Goh, Gabriel and Agarwal, Sandhini and Sastry, Girish and Askell, Amanda and Mishkin, Pamela and Clark, Jack and others},
  booktitle={International conference on machine learning},
  pages={8748--8763},
  year={2021},
  organization={PmLR}
}

@inproceedings{jiang2026sra,
  title={Representation Alignment for Diffusion Transformers without External Components},
  author={Jiang, Dengyang and Wang, Mengmeng and Li, Liuzhuozheng and Zhang, Lei and Wang, Haoyu and Wei, Wei and Dai, Guang and Zhang, Yanning and Wang, Jingdong},
  booktitle={International Conference on Learning Representations},
  editor={Vondrick, Carl and Hariharan, Bharath and Raffel, Colin and Pinto, Lerrel and Yang, Diyi and Faust, Aleksandra},
  pages={18884--18915},
  volume={2026},
  year={2026}
}

@misc{xiang2026conditioningresiduals,
      title={Conditioning Residuals for Diffusion Models via Representation Feedback}, 
      author={Weilai Xiang and Hongyu Yang and Di Huang and Yunhong Wang},
      year={2026},
      eprint={2505.10999},
      archivePrefix={arXiv},
      primaryClass={cs.CV},
      url={https://arxiv.org/abs/2505.10999}, 
}

@article{dai2026probing,
  title={Probing Diffusion Denoising Dynamics for Contrastive Representation Learning},
  author={Dai, Yasong and Hayder, Zeeshan and Ahmedt-Aristizabal, David and Li, Hongdong},
  journal={arXiv preprint arXiv:2607.09067},
  year={2026}
}

@inproceedings{li2023mage,
  title={Mage: Masked generative encoder to unify representation learning and image synthesis},
  author={Li, Tianhong and Chang, Huiwen and Mishra, Shlok and Zhang, Han and Katabi, Dina and Krishnan, Dilip},
  booktitle={Proceedings of the IEEE/CVF conference on computer vision and pattern recognition},
  pages={2142--2152},
  year={2023}
}

@article{xiang2025ddae++,
  title={DDAE++: enhancing diffusion models towards unified generative and discriminative learning},
  author={Xiang, Weilai and Yang, Hongyu and Huang, Di and Wang, Yunhong},
  journal={arXiv preprint arXiv:2505.10999},
  year={2025}
}
